\documentclass[journal]{IEEEtran}
\usepackage{graphicx}

\ifCLASSINFOpdf
\else
\fi

\begin{document}

\title{Feature Interaction based Neural Network for Click-Through Rate Prediction}
%
%
%

\author{Dafang Zou,
        Leiming Zhang,
        Jiafa Mao,
        Weiguo Sheng,~\IEEEmembership{Member,~IEEE}
}

%
%

\maketitle

\begin{abstract}
Click-Through Rate (CTR) prediction is one of the most important and challenging in calculating advertisements and recommendation systems. To build a machine learning system with these data, it is important to properly model the interaction among features. However, many current works calculate the feature interactions in a simple way such as inner product and element-wise product. This paper aims to fully utilize the information between features and improve the performance of deep neural networks in the CTR prediction task. In this paper, we propose a Feature Interaction based Neural Network (FINN) which is able to model feature interaction via a 3-dimention relation tensor. FINN provides representations for the feature interactions on the the bottom layer and the non-linearity of neural network in modelling higher-order feature interactions. We evaluate our models on CTR prediction tasks compared with classical baselines and show that our deep FINN model outperforms other state-of-the-art deep models such as PNN and DeepFM. Evaluation results demonstrate that feature interaction contains significant information for better CTR prediction. It also indicates that our models can effectively learn the feature interactions, and achieve better performances in real-world datasets.
\end{abstract}


\IEEEpeerreviewmaketitle

\section{Introduction}

\IEEEPARstart{M}{any} industrial applications scuh as Online advertising~\cite{66,67}, recommender system~\cite{68}, and web search~\cite{69,70} are Cost per Click(CPC) for most of Internet companies. In the CPC advertising system, the RankSore of advertisements is generally determined by the product of bid price and measurement of advertising items~\cite{1}. In advertising systems, click-through rate (CTR) prediction is generally used as a measure of advertising items. 
\begin{equation}
	RankScore=bid\times CTR
	\label{eq:1}
\end{equation}
Therefore, Correctly predicting CTR of ads is a prerequisite for ensuring reve-nue and user experience. The CTR prediction problem is a typical supervised machine learning problem whose goal is to accurately predict the probability of user’s behavior under the premise of a given advertising item, user and query context. The features of this supervised learning problem are denoted as $x$, and the target is denoted as $y$. By collecting online advertisements and click logs, we can obtain a large number of labeled samples ${(x_{i},y_{i})}$ as training data for supervised learning. We use a parametric w to model this probability:
\begin{equation}
CTR=Probalility(click|AD, User, Query)=f(x, w)
\label{eq:2}
\end{equation}
This supervised learning is an optimization problem where the search a proper w minimizes the objective loss function $L(y, f(x, w))$. The negative Log-Likelihood function is generally used as the loss function in the CTR prediction problem. Many classical machine learning models, including logistic regression (LR)~\cite{8}, Bayesian models~\cite{9}, polynomial-2 (Poly2)~\cite{6}, gradient boosting decision tree~\cite{12, 46, 47}, tensor-based models~\cite{57}, and factorization machines (FM)~\cite{13,37,14,15}, have been proposed in this field.

In order to improve the fitting ability of complex relationships, in feature en-gineering, first-order discrete features are often combined in pairs to form higher-order combined features. To build an effective machine learning (ML) model with CTR prediction, it is crucial to model interactions between features. It has been proved in the Kaggle competition that crafting combinatorial features is an effective way for CTR prediction~\cite{7,17,44,29}. For example, in the CTR predic-tion task, the row data has two discrete features: language and type. In order to improve the fitting ability, we can cross feature language=\{Chinese, English\} with type=\{movie, teleplay\} and get a new feature language\_type=\{Chinese\_movie, Chinese\_teleplay, English\_movie, English\_teleplay\}. However, the performance improvement comes with high cost, since it requires a lot of manual feature engineering as well as domain knowledge to crafting effective feature interactions, and the construction process is not universal.

Instead of designing new featues manually,  another way is to apply machine learning algorithms to learn feature interactions from raw data automatically. Factorization machines (FM)~\cite{13}, proposed to solve the automatic feature combination problem via inner product of  feature embedding vectors, which is regarded as one of the most successful embedding models~\cite{22} in this field. Deep neural networks (DNN) have achieved success in image classification~\cite{50,18,27}, natural language processing (NLP)~\cite{51,19} and speech recognition~\cite{20} over the past years. DNN can automatically capture feature representations and dependencies for prediction purpose compared with linear models. As a result, several DNN based methods for CTR prediction has been proposed in this field. One challenge of applying DNN in CTR prediction is sparsity. Most of these data (such as user ID, gender and city etc.) in CTR problem are non-contiguous and discrete, which are typically converted to a set of high-dimensional sparse fea-tures via the one-hot encoding~\cite{3,4,5,6,7}. For those spare data, they firstly need to be converted into dense feature embedding vectors before inputing to the neural networks. Besides the transformation for the raw features, we also need to provide representations for the feature interactions in the neural networks structure because the representing feature interactions is crucial for CTR prediction tasks. Intuitively, we can directly utilize the feature interactions vector to represent interactions of two features. However, the direct representation of two features usually leads to little sufficient value whose parameters cannot obtain adequate training since row data are sparse. Therefore, we could utilize two feature embedding vectors to caculate feature interactions, such as inner ~\cite{13,29} and element-wise product~\cite{21,32}.

In this work, we propose a novel model for CTR prediction named Feature Interaction based Neural Network (FINN), which enhances DNNs by modelling 2-order feature interactions after feature embedding. By employing a new feature interaction operation in neural network modelling, We have improved the ability of neural networks to learn feature cross information. In order to improve performance further, we also deepen the shallow model combining a classical deep neural network(DNN) component, modelling higher-order and nonlinear feature interactions effectively to improve expressiveness. In contrast to traditional methods that simply perform inner product or element-wise on embedding vectors to model feature interaction in the low level, our proposed feature  mechanism encodes more informative feature interactions, greatly facilitating the following deep layers to learn meaningful information. We conduct extensive experiments on two real-world datasets and show that our model out-performs other models such as factorization machine(FM) and the other state-of-the-art deep models such as PNN and DeepFM.

The remainder of the paper is organized as follows. Related works are introduced in Section 2. The model details are described in Section 3. Section 4 exhib-its experiments’ analysis. Finally, we conclude the paper in Section 5.

\begin{figure*}
	\centerline{\includegraphics[width=0.66\textwidth]{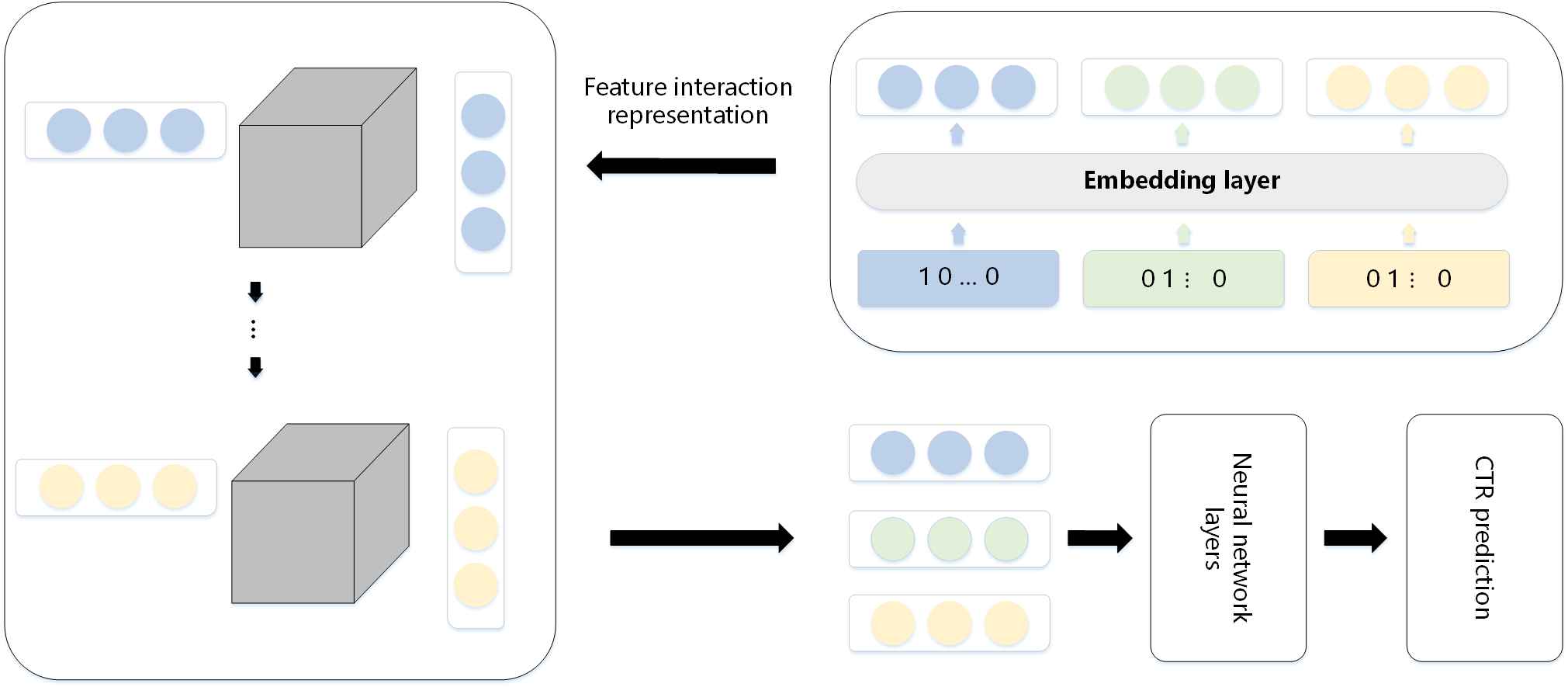}}
	\caption{Architecture of our model. Note: Yellow square denotes input by employing the one-hot encoding; Letter `e' denotes $k$-dimension embedding vector; Symbol `$\times$' denotes element-wise product of two vectors; Letter `w' denotes the weighted accumulation.
	}
	\label{fig_INN}       
\end{figure*}

\section{Related Work}
Many methods have been proposed to process high dimensional sparse data in literature. The section will review word embedding and neural network based methods, which are related to our work.
\subsection{Shallow Methods}
Logistic regression (LR)~\cite{58} is a linear model, which is a widely used model in CTR task. Its corresponding optimization problem has very good properties, which is an unconstrained convex optimization problem with a globally unique optimal solution. It supports large-scale features and can quickly converge to the optimal solution through the commonly used gradient descent method. The interpretability of the logistic regression model is very good. Through the weights corresponding to the features, we can analyze the importance of each feature and their influence on the click rate. However, its expression ability is relatively weak as a linear model, and improve the expression ability of the model through a large number of feature engineering, such as feature combination.

Factorization machine(FM)~\cite{13} proposed to learn the feature interactions using inner product, which is one of the most successful CTR model.
\begin{equation}
\hat{y}_{FM}(x) = w_{0} + \sum_{i=1}^{n}{w_{i}x_{i}} + \sum_{i=1}^{n}{\sum_{j=i+1}^{n}{\langle v_{i},v_{j}\rangle \cdot x_{i}x_{j}}}
\label{eq:3}
\end{equation}
where $w_0 \in R$ is the global bias, $w_i \in R$ is the weight of $i$-th feature, $v_i \in R^k$ is $k$-dimensional vector, and $\langle v_{i}, v_{j}\rangle$ defines inner product. FM performers well on large sparse data, and it has a low time complexity. Compared with the FM model, the Field-aware factorization machine (FFM) ~\cite{14,15} introduces the concept of field. In FM, feature $i$ using the same vector makes interactions with other features, while in FFM, feature i using different vectors makes interactions with features of different fields, which further improves expressiveness of the model. However, FFM was limmited by the need of large memory and cannot easily be applied in real CTR task.

\subsection{Neural Network based Methods}
Deep learning has achieved great success in many research fields such as computer vision~\cite{61,62} and natural language processing~\cite{63}. As a result, many deep learning based CTR models have also been proposed in recent years~\cite{59,60,64}. How to effectively model the feature interactions is the key factor for most of these neural network based models. 

In traditional DNN model, the input of network is usually dense and numberical, while the case of CTR task data whose dimension could over one million after one-hot encoding is not directly applicable. To deal with such an issue, factorization machine supported neural networks (FNN) has also been proposed, which combines an embedding layer pretrained by FM to convert sparse features to a low dimension and dense space and a DNN component to capture high-order feature interactions~\cite{22}. To strengthen model’s capacity of feature interactions, product-based neural network (PNN)~\cite{29} and its extension product-network in network (PIN)~\cite{55} introduces product operations performed on the embedding layer before applying full-connected DNN. Wide \& Deep’s model trains both the shallow component and deep component~\cite{30} at same time. The shallow component based on linear model such LR has the benefits of memorization for low order features while the deep component based on DNN improves the generalization of model. However, features of the linear component used directly for final prediction require manual design to ensure a good result in the Wide \& Deep. Therefore, the DeepFM~\cite{56} tries to replace the linear part of wide \& deep model with FM to learn feature interactions and avoid manual feature engineering and employ the same feature embedding vectors between the shallow component and DNN. Similarly, deepFM jointly train the FM part and the DNN part, which is regarded as one state-of-the-art model. As for neural factorization machines (NFM)~\cite{21}, it is an approach used DNN to improve FM. As mentioned, FM models the same weights of feature interactions over to the CTR result prediction. Supposing the contribution of each feature interactions to CTR prediction result is different, attention neural factorization machines (AFM)~\cite{32} utilized an attention mechanism originated from Neural Machine Translation (NMT) field~\cite{65} to learn the weights of feature interactions. Deep Interest Network (DIN)~\cite{1} designs an a local activation unit structure to adaptively capture user diverse interests from historical behaviors.

\section{Proposed Method}
Our method’s main purpose is  to model the feature interaction representation in a more effective way. To this end, we propose Feature Interaction based Neural Network for CTR prediction tasks.

In this section, we will present the proposed model from the following parts: sparse input layer, embedding layer, Feature-Interaction layer, combination layer, multiple hidden layers and output layer. The architecture of the FINN model as illustrated in Figure 1. The logistic regression part is not shown for clarity purpose. The sparse input layer and embedding layer are the same with DeepFM~\cite{56}, which adopts a sparse representation for input features and embeds the raw feature input into a dense vector. The following Feature-Interaction layer models second order feature interactions on the original embedding. Subsequently, these cross features are concatenated by a combination layer which merges the outputs of Feature-Interaction layer. At last, we feed the cross features into a deep neural network and the network outputs the prediction score.
\subsection{Sparse Input and Embedding layer}
Unlike image classification or speech recognition, the input data in CTR task are usually non-contiguous and categorical. To represent for raw input features, they are normally converted to a high-dimensional sparse features via the one-hot encoding. For example, user\_id = \{001, 002, … \}, goods = \{book, basketball, … \} and gender = \{male, female\}. By employing the one-hot encoding, one input instance can be transformed into:
$${\underbrace{[1,0,0,0,...,0]}_{user\_id=001}}\quad {\underbrace{[0,1,0,0,...,0]}_{goods=book}}\quad
{\underbrace{[0,1]}_{gender=female}}$$
The dimension of above features, especially the user ID and goods type, will become huge after converting. For example, if the number of goods is 550, the dimension of goods feature will increase to 550 after coding, and only one of the 550 values is effective. The sparseness of the coded feature suggest that DNNs is not directly applicable, a solution to utilize the neural network in these case is embed sparse feature into a continuous, dense real-value vector space with a low dimension. The architecture of embedding layer is illustrated in Figure 2. The result of embedding layer is a wide concatenated field embedding vector: 
$$E=[e_1, e_2, …, e_i, …, e_m]$$
where $m$ denotes the number of fields, $e_i\in R^k$ denotes the embedding vector of $i$-th field, and $k$ is the dimension of embedding vector.
\begin{figure}
	\centering
	\includegraphics[width=2.5in]{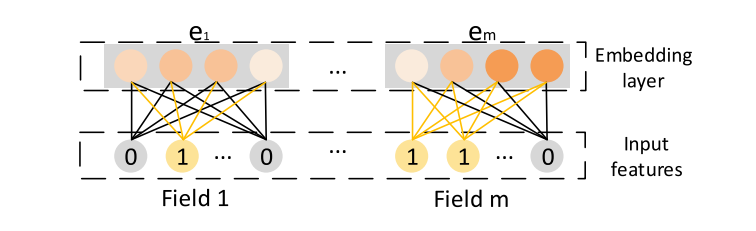}
	\caption{Architecture of embedding layer. The dimension of embedding vector is 4.}
	\label{fig_embedding}       
\end{figure}
\subsection{Feature-Interaction Layer}
To improve prediction accuracy for CTR task, it is useful to provide representations for the feature interactions after raw features embedding layer. The Feature-Interaction layer aims to model the second order feature relations in a precise and effective way. Intuitively, we can directly utilize the feature interactions vector $p_{ij}$ to represent interaction of $i$-th feature and $j$-th feature. The number of the feature interaction vector is $n(n-1)/2$ , where n denotes the number of coded features. However, it is difficult to adequately train the vector $p_{ij}$ in practical application scenarios where data sparsity is widespread. The reason is that the training of each parameter pij requires a large number of samples with non-zero $x_i$ and $x_j$. Because the sample data is inherently sparse, there will be very few samples that satisfy both $x_i$ and $x_j$. Insufficient training samples can easily lead to inaccurate $p_{ij}$ parameters, which will ultimately seriously affect the performance of the model. To deal with such an issue, one of solutions is using embedding vector $x$ to caculate the interaction vector $p$. Inner product and element-wise product are currently the most classical methods for calculate feature interaction. Shallow models such as FM and FFM commonly employ inner product, deep model such as PNN and NFM widely adopt the element-wise product. The terms of inner product and element-wise product can be respectively defined as:
\begin{equation}
f_{inner}(\varepsilon) = \{(v_i \cdot v_j)x_ix_j\}_{(i, j)\in \mathcal{R}_\mathcal{X}}
\label{eq:4}
\end{equation}
\begin{equation}
f_{element-wise}(\varepsilon) = \{(v_i \odot v_j)x_ix_j\}_{(i, j)\in \mathcal{R}_\mathcal{X}}
\label{eq:5}
\end{equation}
where $\mathcal{R}_\mathcal{X} = \{(i,j)\}_{i\in \mathcal{X},j\in \mathcal{X}, j>i}\}$, $v_i$ is the $i$-th embedding vector, $\cdot$ defines the  inner product, and $\odot$ defines the element-wise product. $()_k$ defines $k$-th dimension value of the vector, that is, $(v_i \odot v_j)_k = v_{ik}v_{jk}$. 

A major problem with inner product and element-wise product methods in interaction representation is that they are too simple to effectively calculate the interactions of feature vectors. Therefore, we propose a method to represent feature interaction vector. We take the $i$-th feature vector and the $j$-th feature vector as an example, the feature interaction vector $p_{ij}$ of the two feature vectors can be defined as: 
\begin{equation}
p_{ij}=[p_{ij}^1,...,p_{ij}^u,...,p_{ij}^l ]
\label{eq:6}
\end{equation}
where $p_{ij}^u$ is the $u$-th dimension value of interaction vector, $l$ is the dimension of interaction vector $p_{ij}^u$ can be expressed as:
\begin{equation}
p_{ij}^u=v_i \cdot W_u \cdot v_j^T
\label{eq:7}
\end{equation}
where $W\in R^{k\times k\times l}$ is the 3-dimensional tensor. Each slice $W_{u,u\in\{1,2,...,l\}}$ of the tensor $W$ represents the $i$-th relation matrix. Figure 3 shows the representation of different feature interaction methods. Based on the original embedding $E$, we get the result of the Feature-Interaction layer, which denotes as $\{p_1, ...,p_i, ...,p_n\}$.
\begin{figure}
	\centering
	\includegraphics[width=2.5in]{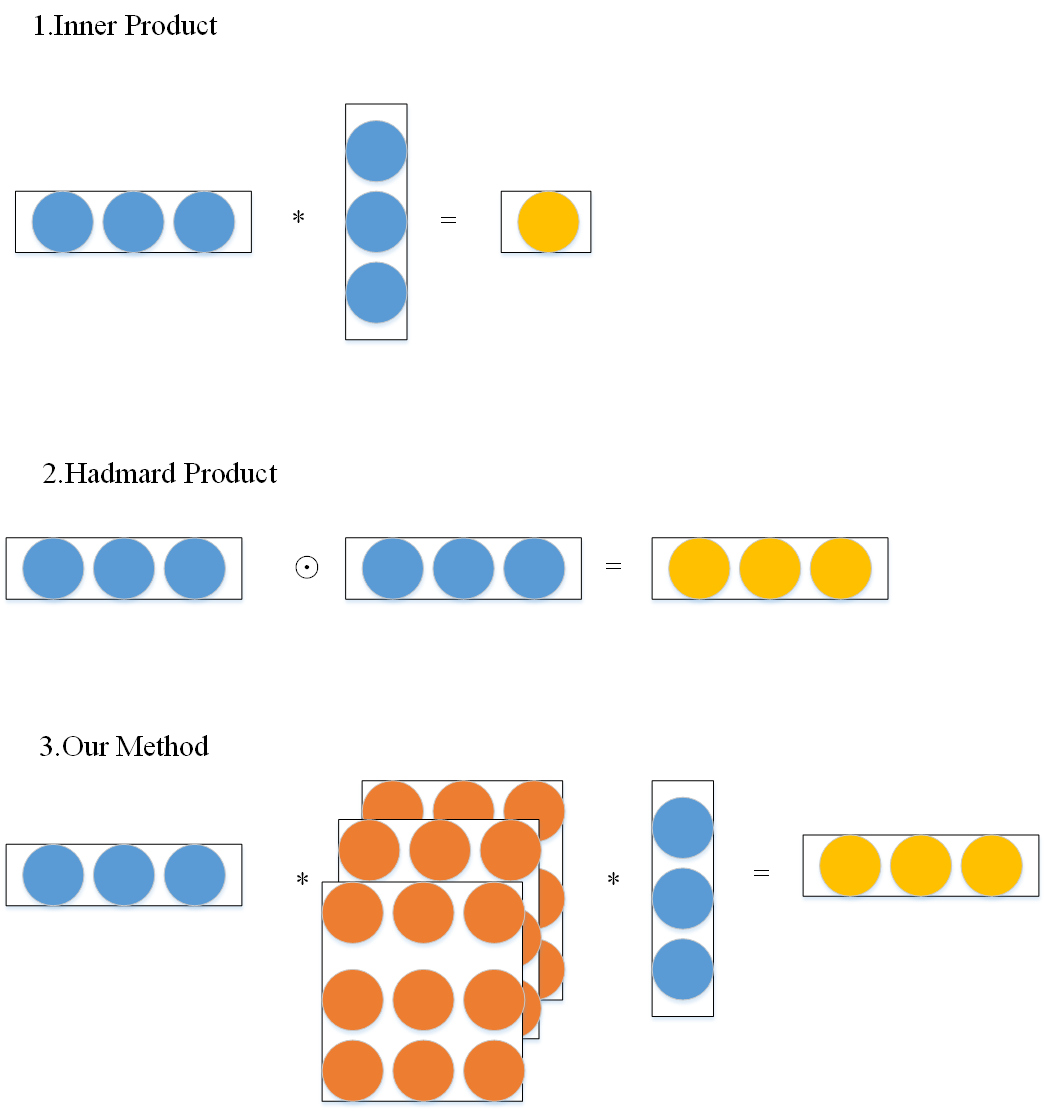}
	\caption{The different methods to calculate feature interaction.}
	\label{fig_interaction}       
\end{figure}
\subsection{Deep Network}
These interaction vectors $p$ are concatenated and then feeds into the deep component which is a feed-forward neural network. The definition of result of combination layer is as follows:
\begin{equation}
F_{concat}({p_1,...,p_i,...,p_n })=[c_1,...,c_i,...,c_k]
\label{eq:8}
\end{equation}
The deep network is used to captures high-order interaction between features and generate the model result. Let $h^{(0)}=[c_1,c_2,...,c_n]$ denotes the inputs of the deep network, where $n$ is the total size of interaction vectors. Formally, the definition of each  full-connected neural network layer is as follow:
\begin{equation}
h^{(l)}=\sigma(W^{(l)}h^{(l-1)}+b^{(l)})
\label{eq:9}
\end{equation}
where $l$ is the layer number of deep network and $\sigma$ is the activation function. $W_{l}\in R^{D_{l+1} \times D_{l}}$, $b^{(l)}$, and $h_{l}\in R^{D_{l}}$ are the model weight, bias and output , respectively, of the $l$-th layer. The deep network is allowed to capture higher-order feature interactions by non-linear activation functions, such as sigmoid, tanh, and ReLU. At last, the output vector of the last neural network layer is generated which is used to caculate the final CTR prediction:
\begin{equation}
y_d=\sigma(W^{|L|+1}h^{|L|} +b^{|L|+1})
\label{eq:10}
\end{equation}
where $|L|$ is the depth of DNN, $\sigma$ is the sigmoid function, defined as $\sigma(x)=1/(1+e^{-x})$.
\subsection{Output Layer and Learning}
To summarize, we show the overall formulation of INN model’s output as: 
\begin{equation}
\hat{y}=\sigma(w_{0} + \sum_{i=1}^{n}{w_{i}x_{i}}+y_d)
\label{eq:11}
\end{equation}
where $y\in (0,1)$ is the value of CTR prediction, $\sigma$ is the sigmoid function, n is the total size of feature, $x$ is a sparse coded input and wi is the $i$-th weight of sparse feature. In the experiments, we evaluate the performance of our method on CTR task and aim to minimize the loss function as follow:
\begin{equation}
loss = \sum_{x\in\mathcal{X}}-y_i(x)\log (\hat{y}_{i}(x))+ (1-y_i(x))\log (1-\hat{y}_{i}(x))
\label{eq:12}
\end{equation}
where $y_i(x)$ is the ground truth of instance $x$, $\hat{y}_{i}$ is the prediction value of CTR, and $\mathcal{X}$ denotes the set of instances for training.

It should be noted that if we remove the neural network part of our model, the relationship dimension $u$ in the relationship tensor is set to 1, and the relationship matrix $W$ is set to 
$
\left[                
\begin{array}{cccc}   
1      & 0      & \cdots & 0      \\
0      & 1      & \cdots & 0      \\
\vdots & \vdots & \ddots & \vdots \\
0      & 0      & \cdots & 1      \\
\end{array}
\right] \in R^{k\times k}
$
, where $k$ is the dimension of feature vector, so our method is equivalent to FM. If we sum each element in vector $c$ and then use a sigmoid function to output a prediction value, we have a shallow CTR model. 

In practice, the size of the training data is usually relatively large when training deep neural networks, especially in CTR task. If it is necessary to calculate the gradient on the entire training data in each iteration during the training process, more computing resources are required. Therefore, Mini-Batch Gradient Descent is often used to train deep neural networks. Learning rate is an important hy-perparameter in neural network optimization. The value of the learning rate $\alpha$ is very critical in the gradient descent methods. If it is too large, it will not converge. If it is too small, the convergence rate is too slow. Therefore, we often use some methods to adaptively adjust the learning rate in practice, such as AdaGrad~\cite{45}, RMSprop, AdaDelta, Adam~\cite{43}, etc. Adam algorithm is a combination of RMSProp and momentum method. In our experiment, we use Adam as the optimizer to learn the parameters, which is denoted as:
\begin{equation}
M_t=\beta_1M_{t-1}+(1-\beta_1)g_t
\label{eq:13}
\end{equation}
\begin{equation}
G_t=\beta_2G_{t-1}+(1-\beta_2)g_t\odot g_t
\label{eq:14}
\end{equation}
\begin{equation}
\Delta\theta_t=-\alpha\frac{M_t/(1-\beta_1^t)}{\sqrt{(G_t/(1-\beta_2^t))}+\varepsilon}
\label{eq:15}
\end{equation}
where, $\beta_1$ and $\beta_2$ are the decay rates of the two moving averages, and the values are usually $\beta_1=0.9$, $\beta_2=0.99$, $\varepsilon$ is a small constant for numerical stability and the value is empirically setted to $10^{-8}$, $M_t$ and $G_t$ is first moment and second moment, $g_t$ is the real gradient at training step $t$.

\subsubsection{Dropout}
While neural networks have achieved great performance, they are also easy to overfit the training data. When training a deep neural network, we can randomly discard a part of neurons (at the same time discard their corresponding connected edges) to avoid overfitting. This method is called the Dropout~\cite{23}. The neurons chosen to be discarded each time are random. For each neuron, there is a probability $p$ to determine whether to keep it. The average number of activated neurons is the original $p$ percent during train, while all neurons can be activated during test. In the INN model, to avoid feature interactions representations overfit the data, we apply dropout in the feature-interaction layer. Specifically, after feture-interaction layer, we randomly drop of concatenated vector with a certain drop-out ratio $p$. Moreover, we also apply dropout in neural networks to prevent the learning from overfitting.

\subsubsection{Batch Normalization}
In the training of deep neural network, the input of a middle layer is the output of the previous neural layer. Therefore, changes in the parameters of the neural layer will cause a large difference in the distribution of its output. From the perspective of machine learning, if the input distribution of a neural layer changes, then its parameters need to be relearned. This phenomenon is called Internal Covariate Shift. To deal with this problem, it is necessary to make the distribution of the input of each neural layer consistent during the training process. The simplest and most direct method is to normalize each neural layer to make its distribution stable. The Batch Normalization (BN)~\cite{24} method is an effective layer-by-layer normalization method that can normalize any intermediate layer in the neural network. In INN, to avoid the update of feature interactions changing the input distribution to neural networks layers, we apply BN on the output of the feature-interaction layer.
\subsection{Difference with Other Neural Network based Methods}
The neural networks based methods can be summarized into the following steps as: $(1)$ An embedding layer maps high-dimensional sparse feature into low-dimensional distributed representations. $(2)$ Several operations are applied on the embedding vectors to get the medial features. $(3)$ A multi-layer perceptron $(MLP)$ is applied  after the fusion of embedding vector to learn nonlinear relations among features.

The fusion of embedding vector is the most intensive part of deep learning model. This layer is the input layer to enter the deep learning model. The quality of embedding fusion will affect the learning of DNN model. There are basically three operations: concatenate~\cite{30,29}, product~\cite{29} or weight sum ~\cite{21,32,1} embedding vectors. We define neural networks based methods uniformly as:
\begin{equation}
\hat{y}=net(fusion(embed(x)))
\label{eq:16}
\end{equation}
FNN simply concatenates the embedding vector together as input to a multilayer neural network, thus lacking the intersection of features. FNN is defined as:
\begin{equation}
\hat{y}_{FNN}=net(concatenate(embed(x)))
\label{eq:17}
\end{equation}
NFM method simply accumulates the feature intersection vectors directly after the feature intersection, which is defined as:
\begin{equation}
\hat{y}_{PNN}=net(concatenate(interaction(embed(x))))
\label{eq:18}
\end{equation}
Wide \& Deep can be considered to combine FNN and Linear method, which is defined as:
\begin{equation}
\hat{y}_{Wide \& Deep} = \hat{y}_{Linear} + \hat{y}_{FNN}
\label{eq:19}
\end{equation}
DeepFM learns the weight of the feature interaction through FM component. Finally, the output of the FM and the output of the Deep part are used as the final result to participate in the fitting of the final target. The FM part and the deep neural network part share the same Embedding layer.
\begin{equation}
\hat{y}_{DeepFM}=\hat{y}_{FM} + \hat{y}_{FNN}
\label{eq:20}
\end{equation}

\begin{table}
	\caption{Dataset statistics}
	\label{tab:1}       
	\setlength{\tabcolsep}{4mm}{
		\begin{tabular}{lllll}
			\hline\noalign{\smallskip}
			Dataset &  \#instance & \#categories & \#fields & pos ratio\\
			\noalign{\smallskip}\hline\noalign{\smallskip}
			Criteo &  $1\times 10^8$ & $1\times 10^6$ &  39 &  0.5\\
			Avazu &  $4\times 10^7$ & $6\times 10^5$ &  24 &  0.17\\
			\noalign{\smallskip}\hline
		\end{tabular}
	}
\end{table}

In contrast to PNN and DeepFM, which directly perform inner product processing after embedding of two features, we use proposed feature interaction operation to learn feature interaction and retain more information of feature interaction, which extended the inner product and is considered as an end-to-end method.
\begin{equation}
\hat{y}_{FINN}=\hat{y}_{Linear} + net(concatenate(FI(embed(x))))
\label{eq:21}
\end{equation}

Finally, the proposed INN can capture high-order latent feature patterns with multi-layer neural network.

\subsection{Performance Comparison}
\begin{table*}[!t]
	\centering
	\caption{Overall performance on the Criteo and Avazu datasets}
	\label{tab:2}       
	\setlength{\tabcolsep}{8mm}{	
		\begin{tabular}{lllll}
			\hline\noalign{\smallskip}
			\multicolumn{1}{l}{Method} &  \multicolumn{2}{l}{Criteo} & \multicolumn{2}{l}{Avazu}\\
			\cline{2-5}\noalign{\smallskip}
			\multicolumn{1}{}{RMSE} & \multicolumn{1}{l}{AUC} & \multicolumn{1}{l}{Log loss} & \multicolumn{1}{l}{AUC} & \multicolumn{1}{l}{Log loss}\\
			\noalign{\smallskip}\hline\noalign{\smallskip}
			LR &   0.7742 & 0.5742 & 0.7545 & 0.3996\\
			FM &   0.7922 & 0.5509 & 0.7765 & 0.3820\\
			FNN &  0.7987 & 0.5431 & 0.7802 & 0.3801\\
			PNN &  0.7994 & 0.5425 & 0.7807 & 0.3797\\
			Wide \& Deep &  0.7986 & 0.5432 & 0.7806 & 0.3800\\
			DeepFM &  0.7986 & 0.5428 & 0.7804 & 0.3797\\
			INN &  0.8020 & 0.5409 & 0.7818 & 0.3785\\
			\noalign{\smallskip}\hline
		\end{tabular}
	}
\end{table*}
\section{Experiments}
In this section, we present our experiments in detail and compare our proposed INN model with related methods, including datasets, baseline methods, evaluation eetricsdata processing, experimental setups, performance comparison, and the analyses of result. In our experiments, the INN model outperforms major state-of-the-art models in the CTR prediction task on two real-world datasets.
\subsection{Dataset}
\subsubsection{Criteo}
Criteo$\footnote{Criteo http://labs.criteo.com/downloads/download-terabyte-click-logs/}$ includes one month of click records with 98 millions of data examples. There are 13 continuous features and 26 categorical ones, and and there is no feature description released. We split the dataset into two parts: “day6-12” for training, and “day13” for test. For numerical features, we discretized them by equalsize buckets. For categorical features, we removed long-tailed data appearing less than 20 times. Nagetive sown sampling is used since the enormous data volume and serious label unbalance (only 3\% samples are positive), and the re-sulting positive sample ratio is about 0.5. After one-hot encoding, the feature space approximates 1M.

\subsubsection{Avazu}
Avazu$\footnote{Avazu http://www.kaggle.com/c/avazu-ctr-prediction}$ includes several days of click-through data which is ordered chronolog-ically. It includes 40 millions click records. For each click instance, there are 24 data fields.We randomly split the public dataset into two parts: 80\% is for train-ing and 20\% is for testing, and remove categories appearing less than 20 times to reduce dimensionality.

\subsection{Baseline Methods}
We compare INN with 6 models in our experiments, which are implemented with TensorFlow and trained with the Adam optimization algorithm.

LR LR is the classical model in CTR task, which treats the recommendation problem as a classification problem and ranks items by predicting the probability of positive samples.

FM FM learns a feature vector for each feature, and the inner product of the two feature vectors is used as feature interactions.

FNN FNN initializes the Embedding layer with the latent vector of FM as the input of the neural network.

PNN The embedding vectors of different features are no longer simple concat-enating, but use product operations to perform pairwise interactions to obtain the interaction information between features more targetedly.

Wide \& Deep The Wide \& Deep model is a hybrid model consisting of a single layer Wide part and a multilayer deep part.

DeepFM DeepFM improves the Wide \& Deep model by replacing the original Wide part with FM.

INN INN is the proposed model of this paper.

\subsection{Evaluation Metrics}
The evaluation metrics are AUC, and log loss.

AUC refers to area under ROC cure, which is a widely used metric in binary classification. AUC is insensitive to the classification threshold and the positive ratio, and it can quantitatively reflect the model performance measured based on the ROC curve. The larger the AUC, the more likely the classifier is to put the true positive samples first, and the better the classification performance.

Log loss can measure the distance between two distributions, which is another widely used metric for binary classification. The lower bound of log loss is 0, indicating the two distributions perfectly match, and a smaller value indicates better performance.

\begin{figure*}
	\includegraphics[width=\textwidth]{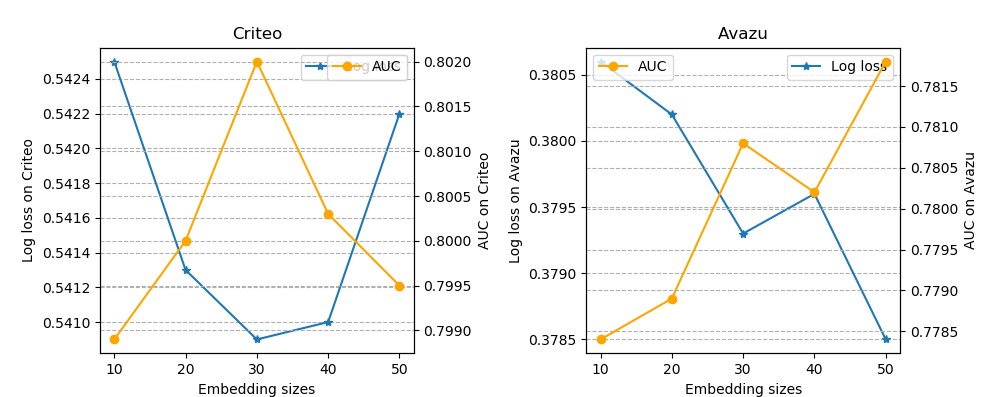}
	\caption{The performance of different embedding sizes on Criteo and Avazu datasets.} \label{fig:embedding}
\end{figure*}
\begin{figure*}
	\includegraphics[width=\textwidth]{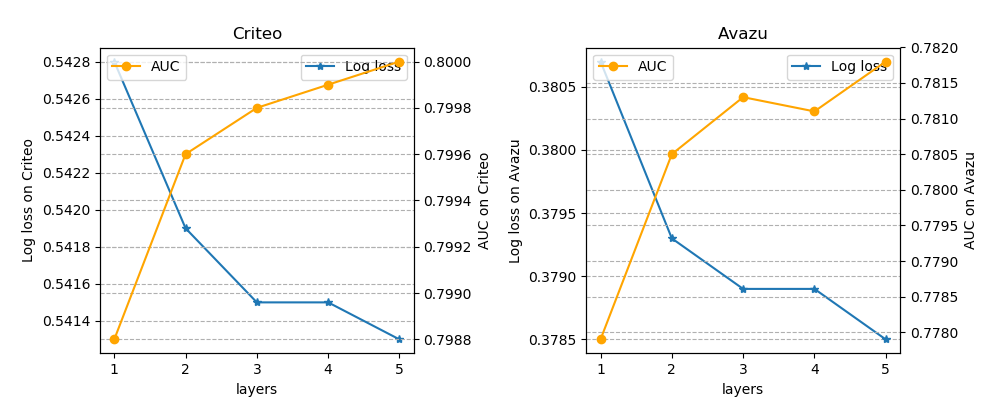}
	\caption{The performance of different number of layers in DNN.} \label{fig:layers}
\end{figure*}
Table~\ref{tab:2} shows the overall performance on Criteo and Avazu datasets, respectively. We implement all the models with Tensorflow$\footnote{TensorFlow: https://www.tensorflow.org/}$ in our experiments. For a fair comparison, the size of embedding vector is set to 30 for Criteo dataset and 50 for Avazu dataset. For the optimization method, we use the Adam~\cite{43} with a mini-batch size of 1000 for Criteo and 500 for Avazu datasets, and the learning rate is set to 0.0001. For all deep models, the depth of layers is set to 5, all activation functions are RELU, the number of neurons per layer is 700 for Criteo dataset and 500 for Avazu dataset. In terms of initialization, we initialize DNN hidden layers with xavier~\cite{71}, and we initialize the embedding vectors from uniform distributions. We conduct our experiments with 2 GTX 1080Ti GPUs.
\begin{table*}[!t]
	\centering
	\caption{Performance comparison between BN and dropout.}
	\label{tab:3}       
	\setlength{\tabcolsep}{8mm}{
	\begin{tabular}{lllll}
		\hline\noalign{\smallskip}
		\multicolumn{1}{l}{Method} &  \multicolumn{2}{l}{Criteo} & \multicolumn{2}{l}{Avazu}\\
		\cline{2-5}\noalign{\smallskip}
		\multicolumn{1}{}{RMSE} & \multicolumn{1}{l}{AUC} & \multicolumn{1}{l}{Log loss} & \multicolumn{1}{l}{AUC} & \multicolumn{1}{l}{Log loss}\\
		\noalign{\smallskip}\hline\noalign{\smallskip}
		FINN &   0.8020 & 0.5409 & 0.7818 & 0.3785\\
		FINN+drop &   0.7983 & 0.5435 & 0.7834 & 0.3780\\
		FINN+BN &  0.8003 & 0.5412 & 0.7796 & 0.3812\\
		\noalign{\smallskip}\hline
	\end{tabular}
}
\end{table*}
\begin{figure*}
	\includegraphics[width=\textwidth]{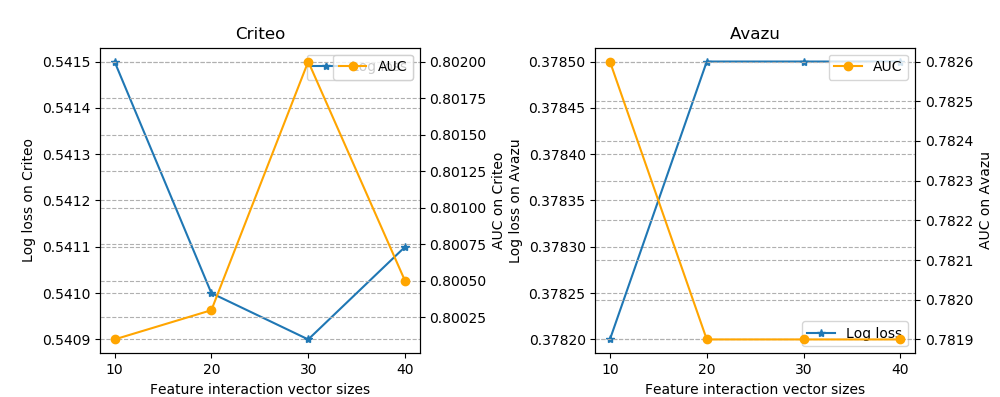}
	\caption{The performance of different number of layers in DNN.} \label{fig:size_interaction}
\end{figure*}
Comparing FM with LR on the AUC and Logloss performance, FM outperforms LR  on two dataset, illustrating the effectiveness of feature interactions. Neural networks based models achieve better performance than FM which only models two-order feature patterns on both datasets, which demonstrates the im-portance of high-order feature interactions. PNN perform better than FNN. A possible reason is the FNN directly e concatenates the feature vectors as the in-put of the neural network makes it hard to explore all possible feature interac-tions. Therefore, we can infer from experimental results that providing representations for the feature interactions on the bottom layer help the neural networks model to gain more expressive power. The im-provement of the DeepFM model mainly aims to improve automatic feature com-bination capabilities of wide part. Unlike PNN, the DeepFM model uses FM for feature interactions. Among all the compared methods, our proposed deep FINN achieves the best result on all metrics on both Criteo and Avazu datasets. The performance of FINN verifie that our proposed feature interaction mechanism to model the feature interaction in bottom DNN for prediction is effective compared with other methods.

\subsection{Parameter Study}
In this subsection, we will conduct some hyper-parameter investigations in our model. We focus on hyper-parameters in the following two components in FINN: the embedding part, the DNN part and feature interaction part. Specifically, we change the following hyper-parameters:(1) the dimension of embeddings; (2) the depth of DNN. Unless specially mentioned in our paper; (3)the dimension feature interaction vector. the default parameter of our network is set as the Section 4.4.

Embedding Part. We change the embedding sizes from 10 to 50 and summarize the experimental results in Figure~\ref{fig:embedding}. We can find some observations as follows. As the dimension is expanded from 10 to 50, our model can obtain a substantial improvement on Avazu dataset. We find 30 are the best for our model when we increase the embedding size on Criteo dataset. Enlarging embedding size indicates increasing the number of parameters in embedding layer and DNN part. We guess that it may be the much more features in Criteo dataset as opposed to Avazu dataset that leads to optimization difficulties.

DNN Part. In deep part, we can change the number of neurons per layer, the depths of DNN, the activation functions. For brevity, we just study the impact different depths in DNN part. As a matter of fact, increasing the number of layers can increase the model complexity. We can observe from Figure~\ref{fig:layers} that increasing number of layers improves model performance at the beginning. However, the performance is degraded if the number of layers keeps increasing. This is be-cause an over-complicated model is easy to overfit. It’s a good choice that the number of hidden layers is set to 5 for Avazu dataset and Criteo dataset.

We find that BN does not improve the performance for FINN. In order to make a choice between the two techniques, we compare models with BN layer and models with 0.5 dropout rate on network hidden layers. Table~\ref{tab:3} illustrates the results. We observe that FINN outperforms FINN+BN for on two metrics on Criteo and Avazu datasets. BN solves internal covariate shift~\cite{24} and accelerates DNN training. However, BN may fail when the input is sparse, because BN relies on the statistics of a mini-batch.

Feature interaction part. In feature layer, we change the feature interaction vector sizes from 10 to 40. As is show in Figure~\ref{fig:size_interaction}, we can find 10 are the best feature interaction vector size on the two metrics on the Criteo and Avazu dataset. In addition, the performance is stable when we increase the feature interaction vector size on Avazu dataset.

\section{Conclusion}
In this work, we proposed a novel neural network model FINN, which brings together the eectiveness of feature interactions machines with the strong representation ability of non-linear neural networks for CTR prediction. The key of FINN’s architecture is the newly proposed feature interaction operation rather than calculating the feature interactions with Hadamard product or inner product, based on which we allow a neural network model to learn more informative feature interactions at the lower level. In order to improve performance further, we combine a classical deep neural network(DNN) component with the shallow model to be a deep model. Extensive experiments on two real-world datasets show that with one hidden layer only, FINN signicantly outperforms LR,  FM, and state-of-the-artdeep learning approaches Wide \& Deep and DeepFM.

\bibliographystyle{IEEEtran}
\bibliography{FINN}

\end{document}